\documentclass[jimaging,article,accept,pdftex,moreauthors]{Definitions/mdpi} 
\usepackage{xfrac}

\firstpage{1} 
\pubvolume{11}
\issuenum{11}
\articlenumber{400}
\pubyear{2025}
\copyrightyear{2025}
\externaleditor{ Jesús Ruiz-Santaquiteria Alegre, Juan Antonio Álvarez García and Harbinder Singh } 
\datereceived{13 October 2025} 
\daterevised{4 November 2025} 
\dateaccepted{5 November 2025} 
\datepublished{8 November 2025 } 
\hreflink{https://doi.org/} 
\usepackage{xurl}
\def\UrlBreaks{%
    \do\/%
    \do\a\do\b\do\c\do\d\do\e\do\f\do\g\do\h\do\i\do\j\do\k\do\l%
    \do\m\do\n\do\o\do\p\do\q\do\r\do\s\do\t\do\u\do\v\do\w\do\x\do\y\do\z%
    \do\A\do\B\do\C\do\D\do\E\do\F\do\G\do\H\do\I\do\J\do\K\do\L%
    \do\M\do\N\do\O\do\P\do\Q\do\R\do\S\do\T\do\U\do\V\do\W\do\X\do\Y\do\Z%
    \do0\do1\do2\do3\do4\do5\do6\do7\do8\do9\do=\do/\do.\do:%
    \do\*\do\-\do\~\do\'\do\"\do\-}

\Title{Benchmarking Compact VLMs for Clip-Level Surveillance Anomaly Detection Under Weak Supervision}

\TitleCitation{Benchmarking Compact VLMs for Clip-Level Surveillance Anomaly Detection Under Weak Supervision}

\Author{{Kirill Borodin} *\orcidA{}, Kirill Kondrashov \orcidB{}, Nikita Vasiliev \orcidC{}, Ksenia Gladkova \orcidD{}, Inna Larina \orcidE{}, Mikhail~Gorodnichev~\orcidF{}  and Grach Mkrtchian *\orcidG{}} 

\AuthorNames{Kirill Borodin, Kirill Kondrashov, Nikita Vasiliev, Ksenia Gladkova, Inna Larina, Mikhail Gorodnichev and Grach Mkrtchian}

\isAPAStyle{%
       \AuthorCitation{Borodin, K., Kondrashov, K., Vasiliev, N., Gladkova, K., Larina, I., Gorodnichev, M., \& Mkrtchian, G.}
         }{%
        \isChicagoStyle{%
        \AuthorCitation{Kirill Borodin, Kirill Kondrashov, Nikita Vasiliev, Ksenia Gladkova, Inna Larina, Mikhail Gorodnichev and Grach Mkrtchian.}
        }{
        \AuthorCitation{{Borodin,} 
 K.; Kondrashov, K.; Vasiliev, N.; Gladkova, K.; Larina, I.; Gorodnichev, M.; Mkrtchian, G.}
        }
}

\address[1]{%
Faculty of Information Technology, Moscow Technical {University} of Communication and Informatics, {Moscow 111024, Russia}; {k.kondrashov@mtuci.ru~(K.K.); n.a.vasilev@edu.mtuci.ru~(N.V.); k.gladkova@mtuci.ru~(K.G.); i.larina@mtuci.ru~(I.L.); m.g.gorodnichev@mtuci.ru~(M.G.)}\\
}

\corres{\hangafter=1 \hangindent=1.05em \hspace{-0.82em}Correspondence: k.n.borodin@mtuci.ru (K.B.);  g.m.mkrtchyan@mtuci.ru (G.M.)}

\abstract{CCTV safety monitoring demands anomaly detectors combine reliable clip‑level accuracy with predictable per‑clip latency despite weak supervision. This work investigates compact vision--language models (VLMs) as practical detectors for this regime. A unified evaluation protocol standardizes preprocessing, prompting, dataset splits, metrics, and runtime settings to compare parameter‑efficiently adapted compact VLMs against training‑free VLM pipelines and weakly supervised baselines. Evaluation spans accuracy, precision, recall, F1, ROC‑AUC, and average per‑clip latency to jointly quantify detection quality and efficiency. With parameter‑efficient adaptation, compact VLMs achieve performance on par with, and in several cases exceeding, established approaches while retaining competitive per‑clip latency. Adaptation further reduces prompt sensitivity, producing more consistent behavior across prompt regimes under the shared protocol. These results show that parameter‑efficient fine‑tuning enables compact VLMs to serve as dependable clip‑level anomaly detectors, yielding a favorable accuracy--efficiency trade‑off within a transparent and consistent experimental setup.}

\keyword{CCTV video analytics; compact vision--language models; parameter-efficient fine-tuning; prompt robustness and design; clip-level anomaly detection}

\begin{document}


\section{Introduction}

\subsection{Context and Relevance}

Closed-circuit television (CCTV) networks are widely deployed across public venues, transit corridors, medical campuses, and industrial operations, where persistent monitoring underpins safety, rapid incident mitigation, and continuity of operations amid constraints such as {limited edge computing capacity} and strict end‑to‑end latency targets~{\cite{hu2023edgebasedvideoanalyticssurvey, abdalla2024videoanomalydetection10, KIM2024105793, xu2023edge}}. Automated video anomaly detection in these environments seeks to surface unusual, safety‑critical events from routine activity, with both missed detections and false alarms imposing measurable burdens on response procedures and resource planning~\citep{abdalla2024videoanomalydetection10, YAO2025101716, ucf_crime}. Consequently, operational deployments prioritize approaches that preserve high accuracy and low, predictable latency despite environmental variability and evolving constraints~\citep{hu2023edgebasedvideoanalyticssurvey, 10.1145/3742794, https://doi.org/10.1002/cpe.6317}.

In operational CCTV settings, camera endpoints generate streams that are handled under bandwidth ceilings, data retention rules, and privacy safeguards, which jointly constrain transmissible content volumes and dictate the timeliness of alarm generation~\citep{YANG2024100204, s22124324, s21093222}.\linebreak These environments prioritize consistent end-to-end latency and seamless integration with incumbent monitoring workflows, thereby shaping method requirements beyond {mere} accuracy considerations~\citep{WAN2022108146}. Within this deployment reality, approaches that preserve dependable detection while satisfying latency budgets are more likely to remain viable for long-term surveillance use~\citep{s21093222, s22124324}.

Automated video anomaly detection processes streaming footage to flag departures from typical behavior patterns, including sudden aggressive interactions, hazardous motions, and object-focused irregularities indicative of elevated risk, while {treating most routine activities as normal variation}~\citep{10084028, ucf_crime, 10.1007/978-3-030-58577-8_20, 9093457, 8578782}. Context determines whether an action is anomalous, creating ambiguity that effective systems must manage, and severe class imbalance further complicates learning because safety-critical incidents are rare relative to everyday scenes~\citep{8578782, ucf_crime}. Contemporary surveys describe two prevailing formulations: learning robust descriptors of normality with deviation scoring, and directly modeling irregular events, with evaluation emphasizing not only accuracy but also timely responsiveness appropriate for surveillance operations~\citep{Raja2023Analysis}.

Deployments in real-world surveillance environments frequently encounter persistent challenges. A major issue is the severe class imbalance, {since} safety-critical incidents occur far less often than routine activities~\citep{10943883, CAETANO2023200236}. Additionally, contextual variations across different cameras, scenes, and locations often lead to domain and distribution shifts, which hinder model generalization and robustness~\citep{Yao_2024_CVPR, 10.5555/3737916.3740772}. Furthermore, strict latency constraints imposed by typical surveillance hardware limit the computational complexity that can be accommodated in real time~\citep{AMINIYEGANEH202495, 10483693}. These factors collectively influence the reliability, scalability, and maintainability of deployed systems, as models that are sensitive to distributional changes or require excessive inference time may reduce operational responsiveness and increase maintenance overhead~\citep{SAMAILA2024127726}. Therefore, practical CCTV-based sensing applications emphasize techniques that maintain high detection accuracy under class imbalance and environmental shifts while adhering to latency requirements, ensuring that algorithmic designs remain consistent with the operational demands of real-world surveillance systems~\citep{10.5555/3737916.3740772}.

These factors highlight the need for a targeted study on automated anomaly detection in CCTV systems, emphasizing both high detection accuracy and {maintaining real-time operational performance}~\citep{s23177442}, thereby establishing an empirical foundation for selecting appropriate methods within realistic application constraints~\citep{AMINIYEGANEH202495}.

\subsection{Research Problem and Associated Challenges}

In this study, video anomaly detection is formulated as a clip-level binary classification problem. Given an input video clip $x$, the system generates a label $y \in {0,1}$ to indicate whether the content is normal or abnormal, without assigning any specific anomaly category~\citep{abdalla2024videoanomalydetection10}. Here, anomalies are context-dependent and are defined as deviations from the usual spatiotemporal behavioral patterns of a particular environment. This contextual definition recognizes that an action considered normal in one location or time may be {deemed} as abnormal in another~\citep{10678181}.

In accordance with the clip-level binary classification framework, the {annotations correspond exclusively to the entire clip, without specifying any temporal boundaries}~\citep{ucf_crime}. This setup represents a weakly supervised learning scenario when compared to segment-level annotation schemes~\citep{10.1109/TCSVT.2024.3350084}. Maintaining both training and evaluation at the clip level ensures methodological consistency~\citep{SAMAILA2024127726} and {avoids assumptions about} the precise temporal localization of events within individual clips~\citep{Yao_2024_CVPR}.

{Given the binary classification setting under weak clip-level supervision}~\citep{s23115024}, the evaluation is performed at the clip level using accuracy, precision, recall, F1 score, and the Area Under the Receiver Operating Characteristic Curve (ROC-AUC). This design ensures consistency with the supervision signal and enables fair comparison across different methods~\citep{Yao_2024_CVPR}. Considering the class imbalance, where abnormal clips are substantially fewer than normal ones, this comprehensive set of metrics is reported instead of relying on a single indicator~\citep{10483693}. Detailed computation procedures are presented later for full reproducibility (see Section~\ref{sec:materials_metrics}). In addition, latency is measured as the average inference time per clip. The hardware specifications and runtime configurations are described in the corresponding section to facilitate transparent and equitable comparison.

In the clip-level weakly supervised context, several challenges hinder accurate anomaly detection. First, context dependence arises when an action may appear normal in one location or time but abnormal in another, leading to ambiguity in interpretation~\citep{10678181}. Second, distribution shifts occur across different cameras, scenes, and time periods, reducing the ability of models to generalize effectively~\citep{10.5555/3737916.3740772}. Third, severe class imbalance is typically present, as abnormal clips are far less frequent than normal ones, which can destabilize training and distort error distributions~\citep{ucf_crime}. Moreover, limited temporal supervision at the clip level can mask short or spatially localized anomalies~\citep{10.1109/TIP.2021.3072863}. Finally, the reporting of average per-clip inference time is essential to properly contextualize performance in relation to the evaluation metrics and practical deployment considerations~\citep{10483693}.

The study is conducted using fixed-length video clips that contain clip-level annotations without temporal localization within each clip~\citep{ucf_crime}. The analysis focuses on binary classification rather than category-specific recognition. The experimental setting is non-adversarial and does not account for intentional attempts to alter or falsify visual content. Advanced capabilities such as identity tracking, person re-identification, or extended multi-camera temporal reasoning are excluded from the present scope~\citep{s23115024}. The evaluation strictly follows the metrics and latency indicators described earlier, maintaining consistent preprocessing and runtime configurations~\citep{9093457}. Comprehensive reporting of implementation details and procedures is provided in the methodology Section~\ref{sec:materials}.

\subsection{Field Snapshot}

Recent video anomaly detection {focuses} on five main approaches~\citep{abdalla2024videoanomalydetection10}: weakly supervised multiple-instance learning~\citep{Sun2020SceneAware, 10.1109/TCSVT.2024.3350084}, vision-language models~\citep{ZANELLA2024104163}, transformer-based temporal modeling~\citep{majhi2024oectst}, self-supervised or contrastive learning~\citep{Hojjati_2024}, and reconstruction- or prediction-based methods~\citep{10027694}. 

\textbf{Weakly supervised multiple instance learning.} Under weak supervision, video anomaly detection largely follows multiple instance learning (MIL), which treats each video as a bag of temporal segments and optimizes snippet-level scores via ranking-style objectives using only video-level labels~\citep{Lv2023Unbiased, Tian2021WeaklySupervised}. Representative methods such as Robust Temporal Feature Magnitude Learning (RTFM)~\citep{Tian2021WeaklySupervised} and transformer-based multi-sequence learning~\citep{Li2022SelfTrainingML} strengthen MIL by maximizing separability among top-k snippets or by learning sequence-level rankings, achieving strong results on UCF-Crime~\citep{ucf_crime}, ShanghaiTech~\citep{shanghai_tech}, and XD-Violence~\citep{10.1007/978-3-030-58577-8_20}. Nonetheless, standard MIL training tends to bias detectors toward the most salient patterns and context shortcuts, causing false alarms on background-correlated cues and overlooking subtle or context-dependent anomalies~\citep{Lv2023Unbiased, abdalla2024videoanomalydetection10}. Recent advances address these issues with prompt-enhanced MIL, which injects semantic priors through learnable prompts to enrich features and sharpen event boundaries~\citep{chen2024prompt}, alongside prompt-based context modeling that improves subclass discriminability at low computational cost~\citep{pu2024learning}. In parallel, normality-guided designs leverage text or visual prompts anchored in normal patterns~\citep{yang2024text} to yield more reliable pseudo-labels~\citep{feng2021mist} and suppress background dominance, improving spatio-temporal localization under weak labels~\citep{pu2024learning}. Despite these improvements, MIL-based Weakly Supervised Video Anomaly Detection(WSVAD) remains constrained by coarse video-level supervision~\citep{Lv2023Unbiased} and a tendency to prioritize easily detected anomalies over boundary cases, motivating richer priors and tighter spatio-temporal grounding in future work~\citep{Tian2021WeaklySupervised}.

\textbf{VLM integration.} Vision-language model integration has introduced training-free and explainable pipelines for video anomaly detection by {pairing VLM captioners with large language models (LLMs) or prompt-driven reasoning components}, as shown in Language-based Video Anomaly Detection (LAVAD)~\citep{Zanella2024HarnessingLL}, Verbalized learning framework (VERA)~\citep{Ye2025VERA}, and AnomalyRuler~\citep{Yang2024FollowTheRules}. LAVAD generates frame-level captions with a VLM and uses an LLM to temporally aggregate and score anomalies without any task-specific training~\citep{Zanella2024HarnessingLL}, demonstrating competitive zero-training performance on UCF-Crime~\citep{ucf_crime} and XD-Violence~\citep{10.1007/978-3-030-58577-8_20}. VERA formalizes verbalized learning by optimizing guiding questions that steer a pretrained VLM to output segment-level scores and human-readable rationales, avoiding parameter updates yet improving detectability and interpretability~\citep{Ye2025VERA}. AnomalyRuler follows an induce-then-deduce scheme: it summarizes normality rules from few-shot normal references via captions, then applies the induced rules to detect anomalies with added smoothing and robustness strategies~\citep{Yang2024FollowTheRules}. Despite these advances, most VLM pipelines still rely on multi-stage components (captioning, reasoning, and post-processing) that increase compute and engineering overhead, and their reliance on general-purpose models can miss surveillance-specific context unless adapted with domain prompts, retrieval, or calibration, as noted in recent surveys and hybrid designs~\citep{Li2024VLAVAD}.

\textbf{Transformer-Based Temporal Modeling.} Transformer-based temporal models aim to capture long-range dependencies in surveillance footage to better localize both brief and prolonged anomalies under weak supervision. Outlier-Embedded Cross Temporal Scale Transformer (OT-CTST)~\citep{majhi2024oectst} exemplifies this trend by embedding anomaly aware temporal positions and using a cross-temporal scale transformer to model correlations across multi-scale features, improving detection of short and long events on UCF-Crime and XD-Violence. Complementing this, Dynamic Erasing Networks~\citep{Zhang2025DENetATM} introduce adaptive temporal modeling that selects per-video temporal scales and progressively erases prominent abnormal segments to reveal subtle ones that standard MIL often overlooks. Transformer Encoded Feature Video Anomaly Detection (TEF-VAD)~\citep{10.1007/978-981-96-2641-0_8} illustrates attention-only designs that employ multi-head attention to enhance feature discrimination within MIL, demonstrating gains on UCF-Crime~\citep{ucf_crime} and ShanghaiTech~\citep{shanghai_tech}. Nonetheless, these transformer pipelines can be computationally heavy for long sequences due to self-attention's quadratic complexity~\citep{DumanKeles2023SelfAttentionComplexity}, motivating scale-aware attention~\citep{Acharya2025StarAttention}, sparsification~\citep{lou2024sparser}, or hierarchical token reduction~\citep{Rao2021DynamicViT} to {improve coverage and transferability.}

\textls[-15]{\textbf{Self-Supervised and Contrastive Learning.} Self-supervised and contrastive methods learn normality from unlabeled video by designing pretext tasks or consistency objectives that shape robust spatio-temporal representations without explicit anomaly labels~\citep{abdalla2024videoanomalydetection10}. Dynamic self-supervised network (DSS-Net)~\citep{10.1109/TMM.2023.3292596} exemplifies this direction by synthesizing spatial and temporal pseudo-abnormal samples to drive dynamic self-supervised training that separates normal and abnormal feature distributions across benchmarks. Contrastive clustering methods such as Cluster Attention Contrast~\citep{10.1145/3394171.3413529} partition normal behaviors into subcategory clusters and maximize agreement within clusters to better cover diverse normal patterns and reduce false positives on atypical but normal events. Hierarchical Semantic Contrast~\citep{Sun2023HSC} integrates object-level and scene-level cues with multi-level contrastive learning to enforce compactness within semantic classes and separability across classes, improving scene awareness and discrimination on standard VAD datasets. Despite these gains, pseudo-anomaly generation~\citep{Rai_2024_CVPRW} can be brittle and may not span the breadth of real-world anomalies, prompting recent work on more generic spatio-temporal pseudo-anomalies using diffusion-guided inpainting and flow perturbations to improve coverage and transferability.}

\textbf{Reconstruction-Based Detection Systems.} Reconstruction-based systems have advanced beyond classic autoencoders by incorporating adversarial training and teacher- student distillation~\citep{CROITORU2024104074} to improve robustness and speed in surveillance scenarios~\citep{10027694}. Multi-scale adversarial distillation learns from strong object-level teachers and transfers their knowledge into a lightweight student via adversarial discriminators~\citep{CROITORU2024104074}, yielding 28--62× speedups with minor accuracy loss on standard VAD benchmarks. Diffusion-based variants further propose reconstruction-free scoring by directly inferring a sample's diffusion latent and evaluating its prior likelihood~\citep{sakai2025reconstructionfree}, achieving state-of-the-art speed-AUC trade-offs without iterative denoising reconstructions. Yet a core limitation persists: high-capacity reconstruction models can over-generalize and faithfully reconstruct abnormal content, compressing the gap between normal and anomalous reconstruction errors and degrading discrimination, as documented and addressed by perturbation-tested and hybrid recon/prediction designs~\citep{10027694}.

\textls[-15]{\textbf{Multimodal Integration and Feature Fusion.} Multimodal integration and feature fusion leverage complementary cues from RGB appearance, optical flow motion, audio events, and even text prompts to improve anomaly localization and robustness in surveillance settings~\citep{Sun2024Multimodal, Shin2025Multimodal}. STPrompt exemplifies prompt-guided fusion by learning spatio-temporal prompt embeddings aligned with video patches to highlight local anomalous regions under weak labels, transferring knowledge from vision-language pretraining for better spatial grounding and temporal consistency~\citep{Wu2024STPrompt}. Beyond prompts, multimodal attention-enhanced fusion combines RGB, flow, and audio streams with dedicated attention modules per modality to capture cross-modal complementarities and reduce missed detections when a single modality is ambiguous or noisy. However, many fusion pipelines still apply equal or static per-modality weighting, which can underperform when anomaly evidence shifts across modalities; recent weakly supervised studies argue for adaptive fusion that reweights modalities by context to better handle motion occlusions, blur, or audio-dominant events~\citep{Shin2025Multimodal}.}

\textls[-15]{The field is converging toward semantic understanding and multimodal reasoning while emphasizing efficiency for real-time surveillance deployment~\citep{Sun2024Multimodal, abdalla2024videoanomalydetection10}. Persistent gaps include generalization across diverse scenes and datasets, robust handling of contextual dependencies over long temporal spans, and interpretable anomaly explanations suitable for operational use~\citep{abdalla2024videoanomalydetection10, wang2025unveilingunseencomprehensivesurvey}. These gaps motivate unified, efficient models that fuse visual and semantic cues to sustain strong detection across varied surveillance contexts and constraints~\citep{Wu2024VADReview, Jebur2024ScalableVAD, Ding2024QuoVadisAD}.}

\subsection{Gap and Rationale}

Despite active progress in vision and language modeling, a controlled cross-model benchmark focused on small VLMs for the target task under a shared protocol is still missing, which limits fair side-by-side comparison and makes it harder to draw cumulative conclusions across studies. Using common dataset splits and uniform scoring is essential so that observed differences reflect model capabilities rather than setup noise, providing a reliable basis to attribute outcomes to the models instead of configuration variance~\citep{chen2024rightwaylvleval}.

There is limited evidence on when parameter-efficient fine-tuning (PEFT), such as Low-Rank Adaptation (LoRA)~\citep{hu2022lowrank}, materially improves small VLMs for this task compared with training-free use~\citep{zanella2024low}, so explicitly testing both regimes is necessary to inform adaptation choices. Prior comparisons often mix architectural differences with divergent prompting practices, which obscures the source of gains; a systematic per-model prompting evaluation clarifies attribution without enforcing a single schema~\citep{chen2024rightwaylvleval}. This focus isolates the effect of adaptation and prompting on measured performance while keeping the benchmark centered on small models under a shared protocol.

Under many existing VLM setups, multi-stage {caption-to-reasoning-to-postprocess} pipelines~\citep{Zanella2024HarnessingLL, Ye2025VERA, Yang2024FollowTheRules} add engineering overhead and complicate integration, which motivates assessing streamlined alternatives that preserve accuracy while reducing complexity. To improve comparability and replication, the benchmark adopts standardized metrics with transparent disclosure of hardware, runtime configuration, and inference settings, so that observed differences are attributable to model behavior rather than undocumented implementation choices.

\subsection{Contribution}

This work introduces a unified, reproducible benchmarking protocol for small vision-language models in video anomaly detection that standardizes clip-level metrics, latency reporting, preprocessing, and runtime settings to enable fair cross-model comparison under surveillance constraints. Within this shared protocol, a consistent prompting schema is specified across instruction-only, zero-shot definitions, few-shot, and reasoning-augmented prompts to disentangle model capability from prompt design and to setup subsequent cross-model and prompt-sensitivity analyses.

Building on the shared protocol, the study systematically evaluates parameter-efficient fine-tuning via LoRA~\citep{hu2022lowrank} against training-free use on compact vision-language models, isolating adaptation effects under consistent metrics, preprocessing, runtime settings, and prompting regimes. The evaluation characterizes how LoRA~\citep{hu2022lowrank} interacts with prompt design and model capacity within a unified pipeline, attributing performance differences to adaptation rather than configuration variance.

\section{Materials and Methods}\label{sec:materials}

\begin{figure}[H]
\includegraphics[width=13.8cm]{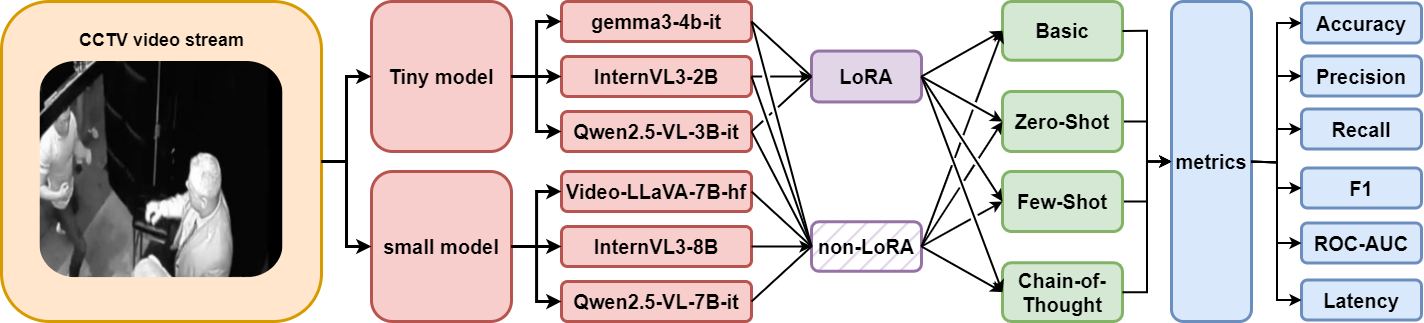}
\caption{{{Overview} of the proposed multimodal inference pipeline for video anomaly detection, integrating LoRA-fine-tuned VLMs with prompting strategies for enhanced accuracy and efficiency.}\label{fig:methodology}}
\end{figure}

This section outlines the methodology employed for detecting deviant behavior in surveillance videos using vision-language models, emphasizing parameter-efficient fine-tuning and prompting strategies to achieve robust anomaly detection. The proposed inference pipeline processes CCTV video streams by extracting frames and applying lightweight and base VLMs, enhanced via LoRA adaptation and different types of prompting, as illustrated in Figure~\ref{fig:methodology}. Evaluations were conducted on the UCF-Crime dataset~\cite{ucf_crime}, measuring accuracy, precision, recall, F1-score, \texttt{ROC-AUC}, and latency to assess performance in image-based security applications.

\subsection{Data}

UCF-Crime~\citep{ucf_crime} is the sole dataset used in this study and serves as a standard benchmark for video anomaly detection; the task is framed as clip-level binary classification (normal vs. abnormal) with no temporal boundaries inside clips to keep the supervision and evaluation units aligned. We adopt the official train/test split without any filtering or exclusions, do not create a validation subset, and report results only on the official test set to maintain strict protocol comparability and reproducibility. {In total, the dataset contains 1900 clips, with 290 reserved for testing (150 normal, 140 abnormal) and 1610 used for fine-tuning (800 normal, 810 abnormal).}

\textls[-15]{We convert each UCF-Crime~\citep{ucf_crime} video into RGB clips using uniform temporal sampling with a variable number of frames: for videos under 60 s, frames are sampled at 0.5 fps across the full duration, while for longer videos, 32 frames are uniformly sampled over the entire timeline; no audio or optical flow is used. Spatial resizing and normalization strictly follow each model's official preprocessing pipeline, with no additional data augmentation, and the exact same sampling and preprocessing are applied during training, fine-tuning, and evaluation to avoid distribution drift. Each clip inherits the source video's binary label without temporal annotations, consistent with the weakly supervised design of UCF-Crime~\citep{ucf_crime}.}

UCF-Crime~\citep{ucf_crime} was selected as the sole benchmark because it provides long, untrimmed CCTV videos across 13 crime categories and is the community standard for weakly supervised anomaly detection, aligning with a clip-level binary protocol and supporting reproducible comparisons. ShanghaiTech~\citep{shanghai_tech} was excluded because it targets frame-level localization in short campus scenes with pixel-level masks and general crowd anomalies, which is outside the scope of clip-level crime detection considered here. XD-Violence~\citep{10.1007/978-3-030-58577-8_20} was excluded due to domain and metric incompatibility: it aggregates heterogeneous, often edited web and movie footage, which conflicts with a CCTV-focused, clip-level binary setup. Consolidating on UCF-Crime~\citep{ucf_crime} thus preserves task, domain, and evaluation consistency while following established surveillance protocols

\subsection{Models}

This study evaluates two capacity tiers to contextualize performance by model size: tiny models (2-4B parameters: \texttt{InternVL3-2B}~\citep{zhu2025internvl3}, \texttt{Qwen2.5-VL-3B}~\citep{bai2025qwen2_5_vl}, \texttt{Gemma-3-4B}~\citep{gemma3_2025}) and small models (7-8B parameters: \texttt{Qwen2.5-VL-7B}~\citep{bai2025qwen2_5_vl},  \texttt{InternVL3-8B}~\citep{zhu2025internvl3},\\
\texttt{Video-LLaVA-7B}~\citep{lin2024videollava}); exact model names and parameter counts are reported to support replication, and Gemma is not included in the small tier because there is no official 7-8B release in that family. 

\texttt{InternVL3-2B} and \texttt{InternVL3-8B} pair~\citep{zhu2025internvl3} a vision encoder with a lightweight language backbone tuned for efficient multi-image/video understanding in constrained compute settings. \texttt{Qwen2.5-VL-3B} and \texttt{Qwen2.5-VL-7B} are general-purpose VLMs~\citep{bai2025qwen2_5_vl} with strong instruction-following, providing robust alignment for video-conditioned text outputs under compact and mid-size budgets. \texttt{Gemma-3-4B}~\citep{gemma3_2025} offers a compact vision-language stack suitable for tiny-tier comparisons but lacks an official 7-8B release, which motivates its exclusion from the small tier. \texttt{Video-LLaVA-7B}~\citep{lin2024videollava} extends LLaVA~\citep{liu2023llava} to video by integrating temporal frame encoding, enabling end-to-end video-to-text reasoning at the small-model scale.

All models are initialized from publicly released checkpoints with no architectural modifications and are loaded using their official tokenizers and preprocessing to avoid confounds. Each model consumes RGB frame sequences with a paired text prompt from the shared prompting suite, and predictions are constrained to a single binary character (“0”/“1”), except under the thinking protocol where free-form reasoning is produced but the final line is still the one-digit answer.

\subsection{Prompting Protocols}

The prompting setup uses four modes applied uniformly across all evaluated models: Basic instruction, Zero-shot definition, Few-shot, and Chain-of-Thought, with fixed templates to ensure comparability across architectures, Appendix~\ref{asec:prompts}. These same templates are used unchanged during fine-tuning and evaluation, maintaining protocol parity and avoiding distribution shift due to wording changes.

The Basic instruction prompt consists of two short sentences that request an abnormal versus normal decision for the given video context and explicitly prohibit providing any rationale text, Appendix~\ref{asubsec:basic_prompt}. The Zero-shot definition prompt offers a concise task framing that defines the decision criterion at a high level and likewise disallows explanation, with wording kept identical across all models for consistency, Appendix~\ref{asubsec:zeroshot_prompt}.

\textls[-15]{The Few-shot prompt provides a compact set of labeled exemplars that illustrate inputs and target labels while explicitly instructing the model to keep any internal reasoning hidden for consistency across runs, Appendix~\ref{asubsec:fewshot_prompt}. The Chain-of-Thought prompt enables explicit multi-step reasoning when applicable, reflecting the assumption that structured deliberation may aid difficult cases even if gains are model- and task-dependent, Appendix~\ref{asubsec:cot_prompt}.}

\subsection{Fine-Tuning and Hyperparameters}

Parameter-efficient supervised fine-tuning is conducted with LLaMA Factory~\citep{zheng2024llamafactory} using LoRA adapters (rank $8$)~\citep{hu2022lowrank} applied to all transformer targets, with base weights and tokenizers kept frozen across all evaluated models. Training uses a single epoch over the configured dataset with a maximum sequence length of $16$ tokens, a per-device batch size of $2$, gradient accumulation of $32$, AdamW with learning rate $2 \times 10^{-5}$, cosine scheduler with 0.1 warmup ratio; bf16 precision is enabled. Runs start from publicly released checkpoints with \texttt{trust\_remote\_code} enabled to follow official preprocessing. {Each tiny vision-language model underwent parameter--efficient fine--tuning with LoRA~\citep{hu2022lowrank} for approximately 40 h.}

{The adapter rank is set to 8 to balance adaptation capacity and stability for classification, consistent with analyses showing that LoRA's low‑rank updates are effective for target tasks while resembling full fine‑tuning only at very high ranks~\cite{shuttleworth2025loravsfinetuningillusion}.} 


\subsection{Inference and Evaluation}\label{sec:materials_metrics}

At test time, clips are evaluated in single-clip batches across all models to standardize runtime conditions and isolate per-clip latency and prediction behavior from batching effects. Wall-clock time is measured per clip as a single forward pass for direct comparability across architectures.

Clip-level performance is reported using {\texttt{Accuracy} (Equation~\eqref{eq:acc}), \texttt{Precision}\linebreak \mbox{(Equation~\eqref{eq:pr}),} \texttt{Recall} (Equation~\eqref{eq:rec}), \texttt{F1} (Equation~\eqref{eq:f1}), and \texttt{ROC-AUC} (Equation~\eqref{eq:auc})} to match the binary abnormal versus normal framing; thresholded metrics are computed from predicted scores via a fixed decision threshold ($0.5$), while \texttt{ROC-AUC} is computed from continuous scores without thresholding. \texttt{ROC-AUC} is obtained by sweeping a threshold over the full score range to trace the ROC curve ($\text{TPR}(\text{threshold})$, $\text{FPR}(\text{threshold})${)}, sorting scores in descending order to update cumulative counts, and averaging ranks for \mbox{tied scores.}
\begin{linenomath}
\begin{equation}\label{eq:acc}
\text{Accuracy} = \frac{\text{TP}+\text{TN}}{\text{TP}+ \text{TN} + \text{FP} + \text{FN}}
\end{equation}
\end{linenomath}
\begin{linenomath}
\begin{equation}\label{eq:pr}
\text{Precision} = \frac{\text{TP}}{\text{TP} + \text{FP}}
\end{equation}
\end{linenomath}
\begin{linenomath}
\begin{equation}\label{eq:rec}
\text{Recall} = \frac{\text{TP}}{\text{TP} + \text{FN}}
\end{equation}
\end{linenomath}
\begin{linenomath}
\begin{equation}\label{eq:f1}
\text{F1} = \frac{2\times\text{Precision}\times\text{Recall}}{\text{Precision}+\text{Recall}}
\end{equation}
\end{linenomath}
\begin{linenomath}
\begin{equation}\label{eq:auc}
\text{{ROC-AUC}} = \int^1_0 TPR(FPR) d(FPR)
\end{equation}
\end{linenomath}

Latency is measured per clip as the wall-clock difference between end time and start time of a single forward pass, then summarized across clips by the sample mean and a $95\%$ confidence interval. For a set of per-clip latencies $\{t_i\}^n_{i=1}$, the mean is
\newpage

\phantom{0}\vspace{-12pt}
\begin{linenomath}
\begin{equation}\label{eq:mean}
\bar{t} = \frac{1}{n} \sum^n_{i=1}t_i,
\end{equation}
\end{linenomath}
the sample standard deviation is

\begin{linenomath}
\begin{equation}\label{eq:deviation}
s = \sqrt{\frac{1}{n-1}\sum^n_{i=1}(t_i-\bar{t})^2},
\end{equation}
\end{linenomath}
and the $95\%$ CI around the mean is computed as $\bar{t}\pm z^* \frac{s}{\sqrt{n}}$ with $z^*=1.96$.
\subsection{{Baselines description}}

External baselines include LAVAD~\citep{Zanella2024HarnessingLL}, VERA~\citep{Ye2025VERA},  and PEL4VAD~\citep{pu2024learning}, all executed from their public repositories on the same UCF-Crime~\citep{ucf_crime} split and evaluated with our clip-level metrics to ensure strict protocol parity with our models, with results reported exclusively from these re-runs on our setup rather than taken from prior papers.

\subsection{Runtime}

Inference was performed on a workstation with two  NVIDIA RTX 4060 Ti 16 GB GPUs and an Intel Xeon E5-2699 v3 CPU, while LoRA~\citep{hu2022lowrank} fine-tuning ran on a separate server with two  NVIDIA Tesla A100 40 GB GPUs paired with an AMD EPYC 7742~CPU; these configurations were used consistently across all reported experiments in their \mbox{respective regimes.}

\section{Results}
 
\subsection{Tiny Models}

Table~\ref{tab:tiny_models_ucf_crime} presents clip-level results on the UCF-Crime dataset~\citep{ucf_crime} for three tiny vision-language \texttt{models---gemma-3-4B-it}~\citep{gemma3_2025} \texttt{InternVL3-2B}~\citep{zhu2025internvl3}, and \texttt{Qwen2.5-VL-3B-It}~\citep{bai2025qwen2_5_vl} evaluated under four prompting regimes. The table reports accuracy, precision, recall, F1, ROC-AUC, and inference latency formatted as mean $\pm$ confidence interval for each model-prompt combination. Model names follow their referenced releases, and the metric set aligns with the study's clip-level evaluation and latency reporting conventions. {\textbf{Bold} {numbers denote the best-performing values for each metric across models and prompting types, while} \underline{underlined} {numbers indicate the second-best results.}}

\begin{table}[H]
\caption{Tiny models comparison with different prompt types on UCF-crime dataset.\label{tab:tiny_models_ucf_crime}}
	\begin{adjustwidth}{-\extralength}{0cm}
		\begin{tabularx}{\fulllength}{c c c c c c c c c c}
			\toprule
			\textbf{Model} & \textbf{Prompt} & \textbf{Accuracy} & \textbf{Precision} & \textbf{Recall} & \textbf{F1} & \textbf{ROC-AUC}   & \textbf{Latency} \\ 
			\midrule
            gemma-3-4B-it~\cite{gemma3_2025} & Basic            & 0.7619 & 0.7105 & \underline{{0.9643}} & \underline{0.8182} & 0.7366 & 9.1008  ±  0.3205 \\
            & Zero Shot        & 0.6825 & 0.6402 & \textbf{{0.9786}}    & 0.7740 & 0.6455 & 7.5379  ±  0.9962 \\
             & Few Shot         & \textbf{0.8056} & 0.8473 & 0.7929            & \textbf{0.8192} & \underline{0.8071} & 8.1017  ±  0.3094 \\
             & CoT & 0.7061 & 0.8211 & 0.5865            & 0.6842 & 0.7173 & 15.1717  ±  0.3021 \\
            \midrule
            InternVL3-2B~\cite{zhu2025internvl3} & Basic prompt & 0.5992  &0.8421  &0.3429 &0.4873 &0.6312 & \underline{2.6972  ± 0.0452} \\
             & Zero Shot & 0.5020  &\textbf{1.0000}  &0.1071 &0.1935 &0.5536 & 2.7692  ± 0.0471\\
             &Few Shot & 0.4484  &\textbf{1.0000}  &0.0071 &0.0142 &0.5036 & \textbf{2.4030  ± 0.0363}\\
             & CoT & 0.5357  &0.6211  &0.4214 &0.5021 &0.5500 & 11.3620  ±  1.0064\\
            \midrule
            Qwen2.5-VL-3B-It~\cite{bai2025qwen2_5_vl} & Basic &  \underline{0.7817}  &\underline{0.9570}  &0.6357 &0.7639 &0.8000 & 5.0947  ±  0.5414 \\
             & Zero Shot&\textbf{0.8056}  &0.8824  &0.7500 &0.8108 &\textbf{0.8125}&2.9228  ± 0.2431\\
             & Few Shot  & 0.5000  &\textbf{1.0000}  &0.1000 &0.1818 &0.5500 & 2.9228  ± 0.2431 \\
             & CoT & 0.6056  &0.7041  &0.4964 &0.5823 &0.6187 & 4.2266  ± 0.5818 \\
			\bottomrule
		\end{tabularx}

	\end{adjustwidth}

\end{table}

\begin{figure}[H]
\begin{adjustwidth}{-\extralength}{0cm}
\centering
\includegraphics[width=16cm]{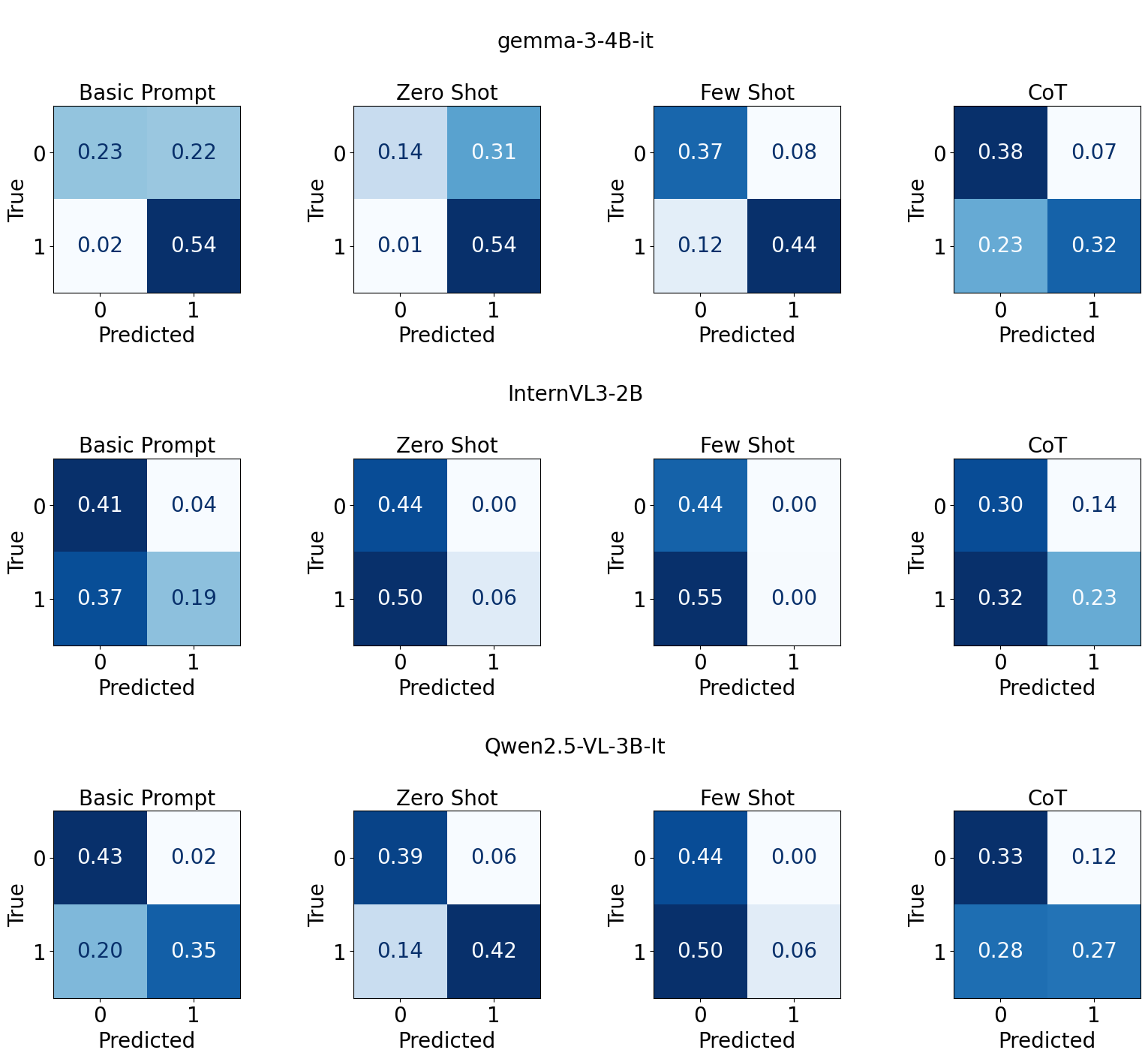}
\end{adjustwidth}
\caption{{{Confusion} matrix for tiny models with different prompt types on UCF-crime dataset.}\label{fig:tiny_models}}
\end{figure}

Figure~\ref{fig:tiny_models} presents a $3\times4$ grid of normalized confusion matrices for three tiny models: \texttt{ {gemma-3-4B-it}}~\citep{gemma3_2025}, \texttt{InternVL3-2B}~\citep{zhu2025internvl3}, and \texttt{Qwen2.5-VL-3B-It}~\citep{bai2025qwen2_5_vl}. Each evaluated under four prompting strategies: Basic, Zero Shot, Few Shot, and CoT. Rows correspond to models and columns to prompting types, with true labels on the y-axis and predicted labels on the x-axis for binary classes 0 and 1. Each panel annotates cell-wise proportions and uses a blue color scale to indicate magnitude for correct and misclassified outcomes within that matrix.

\vspace{-9pt}

\subsection{Small Models}

Table~\ref{tab:small_models_ucf_crime} reports clip-level accuracy, precision, recall, F1, ROC-AUC, and per-clip latency (mean $\pm$ confidence interval) for \texttt{InternVL3-8B}~\citep{zhu2025internvl3}, \texttt{Video-LLaVA-7B-hf}~\citep{lin2024videollava}, and \texttt{Qwen2.5-VL-7B-It}~\cite{bai2025qwen2_5_vl} under Basic, Zero Shot, Few Shot, and CoT prompting on UCF-Crime~\citep{ucf_crime}. The best aggregate classification metrics occur for \texttt{Qwen2.5-VL-7B-It}~\cite{bai2025qwen2_5_vl} with Basic prompting, with the second-best values from \texttt{InternVL3-8B}~\citep{zhu2025internvl3} with Basic prompting. Precision reaches 1.0000 for \texttt{InternVL3-8B}~\citep{zhu2025internvl3} (Few Shot) and \texttt{Qwen2.5-VL-7B-It}~\cite{bai2025qwen2_5_vl} (Few Shot, CoT), and the minimum latency is observed for \texttt{Video-LLaVA-7B-hf}~\citep{lin2024videollava} with Basic prompting. {\textbf{Bold} {highlights the highest score within each metric column, and} \underline{underlining} {marks the next-highest (second-best) value.}}

\begin{figure}[H]
\begin{adjustwidth}{-\extralength}{0cm}
\centering
\includegraphics[width=16cm]{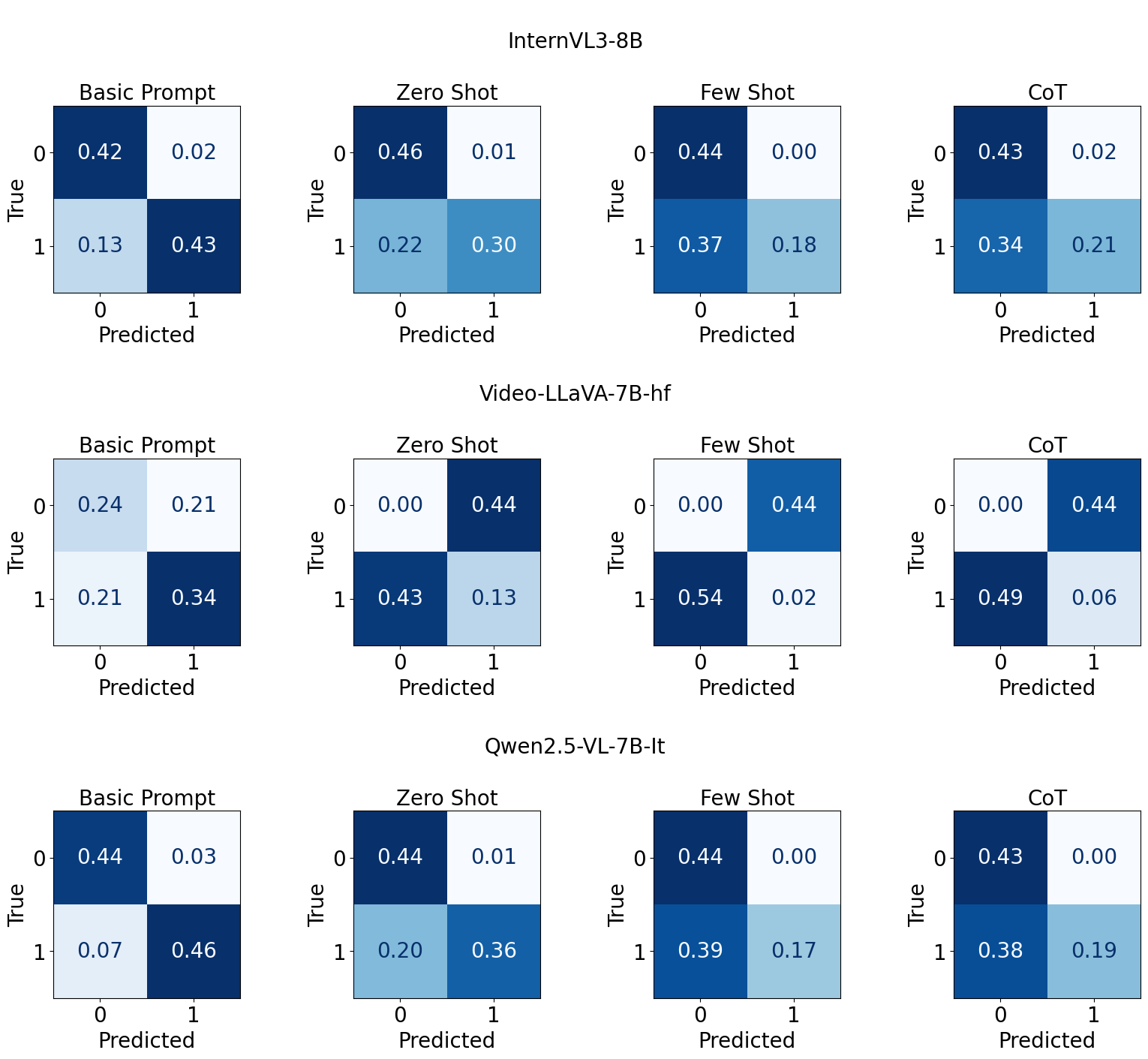}
\end{adjustwidth}
\caption{{{Confusion} matrix for small models with different prompt types on UCF-crime dataset.}\label{fig:small_models}}
\end{figure}

Figure~\ref{fig:small_models} displays a $3\times4$ grid of normalized confusion matrices for three small models---\texttt{InternVL3-8B}~\citep{zhu2025internvl3}, \texttt{Video-LLaVA-7B-hf}~\citep{lin2024videollava}, and \texttt{Qwen2.5-VL-7B-It}~\cite{bai2025qwen2_5_vl} evaluated under four prompting strategies, arranged by rows for models and columns for prompting types. Each panel reports proportions for a binary classification task with ``True'' on the y-axis and ``Predicted'' on the x-axis, annotated with cell values for classes $0$ and $1$. A blue intensity colormap encodes magnitude, with diagonal cells representing correct predictions and off-diagonal cells representing errors for each model-prompt configuration.

\begin{table}[H]
\caption{Small models comparison with different prompt types on UCF-crime dataset.\label{tab:small_models_ucf_crime}}
	\begin{adjustwidth}{-\extralength}{0cm}
		\begin{tabularx}{\fulllength}{c c c c c c c c c c}
			\toprule
			\textbf{Model} & \textbf{Prompt} & \textbf{Accuracy} & \textbf{Precision} & \textbf{Recall} & \textbf{F1} & \textbf{ROC-AUC}   & \textbf{Latency} \\ 
			\midrule
            InternVL3-8B~\cite{zhu2025internvl3} & Basic &\underline{{0.8532}}&0.9558 &\underline{0.7714} &\underline{0.8538 }&\underline{0.863}& 4.8936 $\pm$ 0.1034 \\
             & Zero Shot          & 0.7627 & 0.9595          & 0.5726 & 0.7172 & 0.7729 & 5.0435 ± 0.0978 \\
           & Few Shot           & 0.6270 & \textbf{{1.0000}} & 0.3286 & 0.4946 & 0.6643 & 4.4191 ± 0.1228 \\
           & CoT   & 0.6360  &0.9123  &0.3768 &0.5333 &0.6661 & 15.9177 ± 0.6653 \\
            \midrule
            Video-LLaVA-7B-hf~\cite{lin2024videollava} & Basic            & 0.5794 & 0.6232 & 0.6143 & 0.6187 & 0.5750 & \textbf{2.5132 ± 0.1236} \\
              & Zero Shot        & 0.1315  &0.2238  &0.2302 &0.2270 &0.1196 & 3.3413 ± 0.2293 \\
              & Few Shot         & 0.0159  &0.0345  &0.0286 &0.0312 &0.01438 & 4.1288 ± 0.1180 \\
              & CoT & 0.0635  &0.1250  &0.1143 &0.1194 &0.0571 & 32.7412 ± 0.0109 \\
            \midrule
            Qwen2.5-VL-7B-It~\cite{bai2025qwen2_5_vl} & Basic            & \textbf{0.8990} & 0.9381          & \textbf{0.8667} & \textbf{0.9010} & \textbf{0.9011} & 3.7055 ± 0.9577 \\
            & Zero Shot        & 0.7937 & \underline{0.9783} & 0.6429 & 0.7759 & 0.8125 & \underline{3.3754 ± 0.9720} \\
            & Few Shot         & 0.6111 & \textbf{1.0000}    & 0.3000 & 0.4615 & 0.6500 & 4.1183 ± 0.9790 \\
            & CoT &0.6178  & \textbf{1.0000}  &0.3265 &0.4923 &0.6633 & 17.4928 ± 0.2127 \\
			\bottomrule
		\end{tabularx}
	\end{adjustwidth}

\end{table}
\unskip

\vspace{-12pt}

\subsection{Tiny Models Finetuned with LoRA}

Table~\ref{tab:tiny_models_ucf_crime_lora} summarizes clip-level accuracy, precision, recall, F1, ROC-AUC, and per-clip latency for \texttt{gemma-3-4B-it}~\citep{gemma3_2025}, \texttt{InternVL3-2B}~\citep{zhu2025internvl3}, and \texttt{Qwen2.5-VL-3B-It}~\citep{bai2025qwen2_5_vl} fine-tuned with LoRA~\citep{hu2022lowrank} under Basic, Zero-Shot, Few-Shot, and CoT prompts, with parentheses indicating percentage differences versus corresponding non-LoRA settings. {\textbf{Bold} {is used to emphasize the top metric performance, with} \underline{underlined} values representing the runner-up results under the same evaluation criteria.}

\begin{figure}[H]
\begin{adjustwidth}{-\extralength}{0cm}
\centering
\includegraphics[width=17cm]{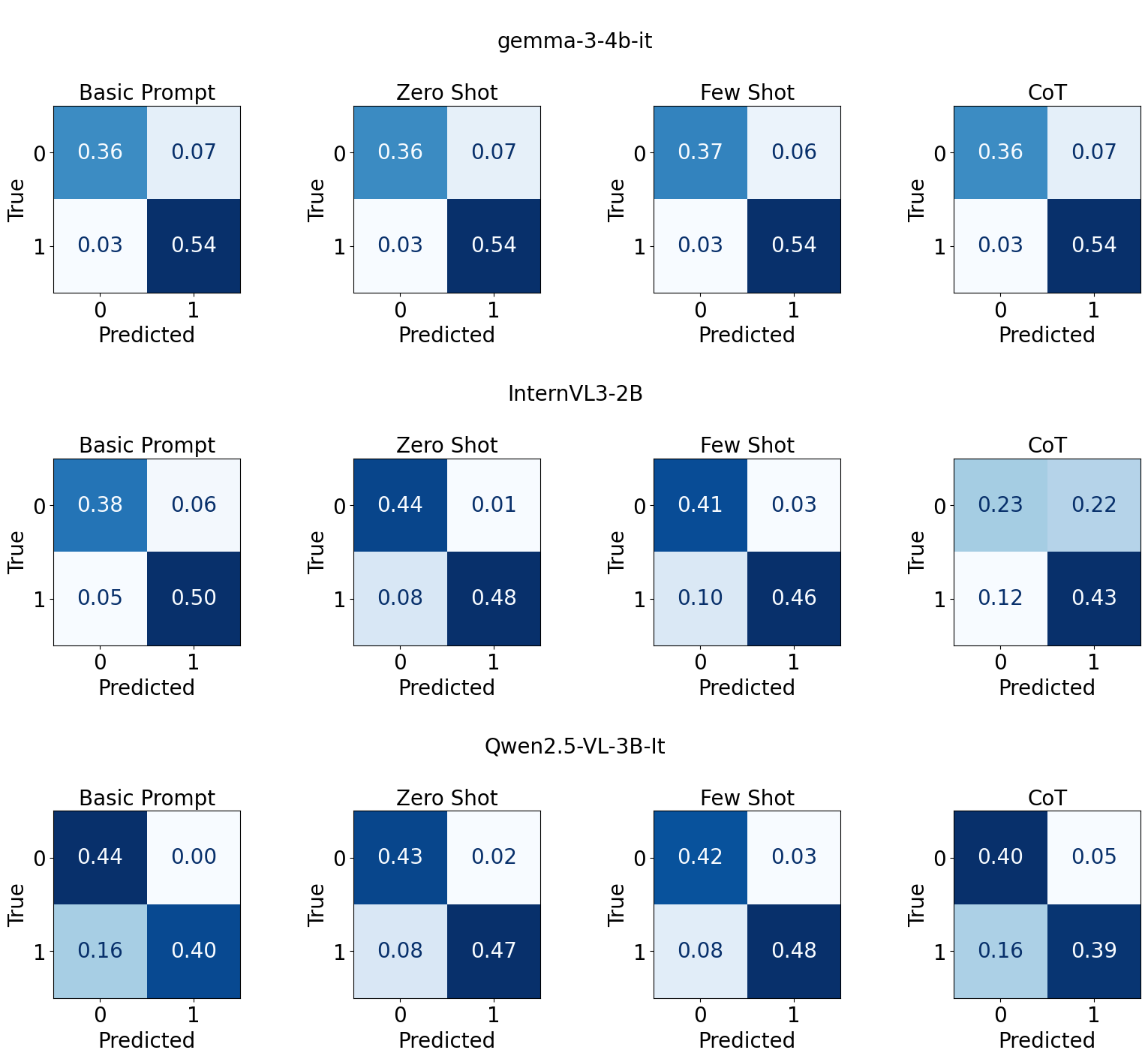}
\end{adjustwidth}
\caption{{{Confusion} matrix for tiny models finetuned with LoRA with different prompt types on UCF-crime dataset.}\label{fig:lora_models}}
\end{figure}

Figure~\ref{fig:lora_models} shows a $3\times4$ grid of normalized confusion matrices for three tiny LoRA-tuned~\citep{hu2022lowrank} evaluated with four different prompts, arranged by rows for models and columns for prompting types. Each panel depicts the proportion of predictions for binary classes with ``True'' on the y-axis and ``Predicted'' on the x-axis, annotated by cell values for correct and incorrect outcomes. A blue colormap encodes magnitude within each matrix, enabling direct visual comparison of per-configuration classification behavior across the four prompting strategies for each model.

\vspace{-12pt}
\unskip

\begin{table}[H]
\scriptsize
\caption{Tiny models finetuned with LoRA comparison with different prompt types on UCF-crime~dataset.\label{tab:tiny_models_ucf_crime_lora}}
	\begin{adjustwidth}{-\extralength}{0cm}
		\begin{tabularx}{\fulllength}{cCcccccccc}
			\toprule
			\textbf{Model} & \textbf{Prompt} & \textbf{Accuracy} & \textbf{Precision} & \textbf{Recall} & \textbf{F1} & \textbf{ROC-AUC}   & \textbf{Latency} \\
			\midrule
            gemma-3-4b-it~\cite{gemma3_2025} & Basic & \underline{{0.90}} (+17.65\%) & 0.87 (+23.04\%) & \textbf{{0.95}} ($-$1.52\%)  & \underline{0.91} (+11.26\%) & 0.89 (+20.83\%) & 8.10 $\pm$ 1.04 ($-$12.35\%) \\
            & Zero Shot & 0.89 (+30.75\%) & 0.87 (\textbf{+36.41\%}) & \underline{0.94} ($-$3.70\%)  & 0.91 (+17.13\%) & 0.89 (+37.32\%) & 8.10 $\pm$ 1.01 (+6.93\%) \\
            & Few Shot & \textbf{0.91} (+13.26\%) & 0.90 (+5.98\%)  & \textbf{0.95} (+19.76\%) & \textbf{0.92} (+12.68\%) & \underline{0.91} (+12.49\%) & 8.37 $\pm$ 0.91 (+3.23\%) \\
            & CoT& \underline{0.90} (+26.95\%) & 0.87 (+6.47\%)  & \textbf{0.95} (+61.91\%) & \underline{0.91} (+33.05\%) & 0.89 (+24.08\%)& 8.77 $\pm$ 1.00 ($-$73.58\%)\\
            \midrule
            InternVL3-2B~\cite{zhu2025internvl3} & Basic prompt & 0.89 (+32.59\%)  &0.89 (+5.85\%)  &0.91 (+62.2\%) &{0.90} (+45.56\%) &0.89 (+28.81\%) &2.91 ± 0.08  (+7.45\%) \\
             &Zero Shot&\textbf{0.91} ({+45\%})  &\underline{0.98} ($-$1.67\%)  &0.86 (+87.49\%) &\textbf{0.92} ({+78.88\%}) &\textbf{0.92} ({+39.8\%}) & 2.77 ± 0.04 (+0.12\%)\\
             &Few Shot&0.87 (\underline{+48.63\%})  &{0.94} ($-$6.89\%)  &0.83 (\underline{+99.14\%}) &0.88 (\underline{+98.38\%}) &0.88 (\underline{+42.68\%}) &\textbf{2.33 ± 0.03} ($-$2.85\%)\\
             &CoT&0.67 (+21.22\%)  &0.67 (+18.27\%)  &0.78 (+45.97\%) &0.72 (+30.44\%) &0.65 (+15.52\%) & 5.09 ± 0.54 (\underline{$-$128.32\%})\\
            \midrule
            Qwen2.5-VL-3B-It~\cite{bai2025qwen2_5_vl}& Basic & 0.84 (+7.53\%) & \textbf{0.99} (+3.46\%) & 0.72 (+13.17\%) & 0.83 (+9.08\%) & 0.86 (+6.90\%) & 2.83 $\pm$ 0.24 (\underline{$-$79.85\%}) \\
             &Zero Shot & \underline{0.90} (+11.77\%) & 0.97 (+9.61\%) & 0.85 (+13.19\%) & 0.90 (+11.52\%) & \underline{0.91} (+11.58\%) & \underline{2.61 $\pm $0.23} ($-$11.96\%)\\
             &Few Shot & \underline{0.90} (\textbf{+79.28\%}) & 0.94 ($-$5.51\%) & 0.86 (\textbf{+763.30\%}) & 0.90 (\textbf{+396.31\%}) & 0.90 (\textbf{+63.71\%}) & 2.77 $\pm$ 0.24 ($-$5.42\%)\\
             &CoT & 0.79 (+30.25\%)& 0.89 (\underline{+26.53\%}) & 0.70 (+42.02\%) & 0.79 (+35.17\%) & 0.80 (+29.13\%)& 2.88 $\pm$ 0.24 ($-$46.76\%) \\
			\bottomrule
		\end{tabularx}
	\end{adjustwidth}

\end{table}

\subsection{Baselines}
Table~\ref{tab:other_models} reports clip-level results on UCF-Crime~\citep{ucf_crime} for four comparative baselines, listing accuracy, precision, recall, F1, ROC-AUC, and per-clip latency with confidence intervals for \texttt{InternVL3-2B}~\citep{zhu2025internvl3} finetuned with LoRA~\citep{hu2022lowrank}, LAVAD~\citep{Zanella2024HarnessingLL}, VERA~\citep{Ye2025VERA}, and PEL4VAD~\citep{pu2024learning}; values are presented as means with the corresponding 95\% confidence intervals following the study's formatting conventions. {\textbf{Bold} {corresponds to the best metric outcome, whereas} \underline{underlined} {entries signify the second-best performance across the compared configurations.}}
\begin{table}[H]
\caption{Comparison with other models on UCF-crime dataset.\label{tab:other_models}}
	\begin{adjustwidth}{-\extralength}{0cm}
		\begin{tabularx}{\fulllength}{Ccccccccc}
			\toprule
			\textbf{Model}  & \textbf{Accuracy} & \textbf{Precision} & \textbf{Recall} & \textbf{F1} & \textbf{ROC-AUC}   & \textbf{Latency} \\ 
			\midrule
            InternVL3-2B-LoRA~\cite{zhu2025internvl3} & \textbf{{0.9127}}  &\textbf{0.9836}  &0.8571 &\textbf{0.9160} &\textbf{0.9196}& \underline{{2.7726} $\pm$ 0.0364} \\
            \midrule
            LAVAD~\citep{Zanella2024HarnessingLL} & 0.7724  &0.7681  &0.7571 &0.7626 &0.7719 & 4.8936$\pm$ 0.1034\\
            VERA~\citep{Ye2025VERA} & 0.6448  &0.5781  &\textbf{0.9786} &0.7268 &0.6560 & 3.7055 $\pm$ 0.9577\\
            PEL4VAD~\citep{pu2024learning} & \underline{0.8724}  &\underline{0.8503}  &\underline{0.8929} &\underline{0.8711} &\underline{0.8731} & \textbf{2.5132 $\pm$ 0.1236}\\
			\bottomrule
		\end{tabularx}
	\end{adjustwidth}

\end{table}

\begin{figure}[H]
\includegraphics[width=13.86cm]{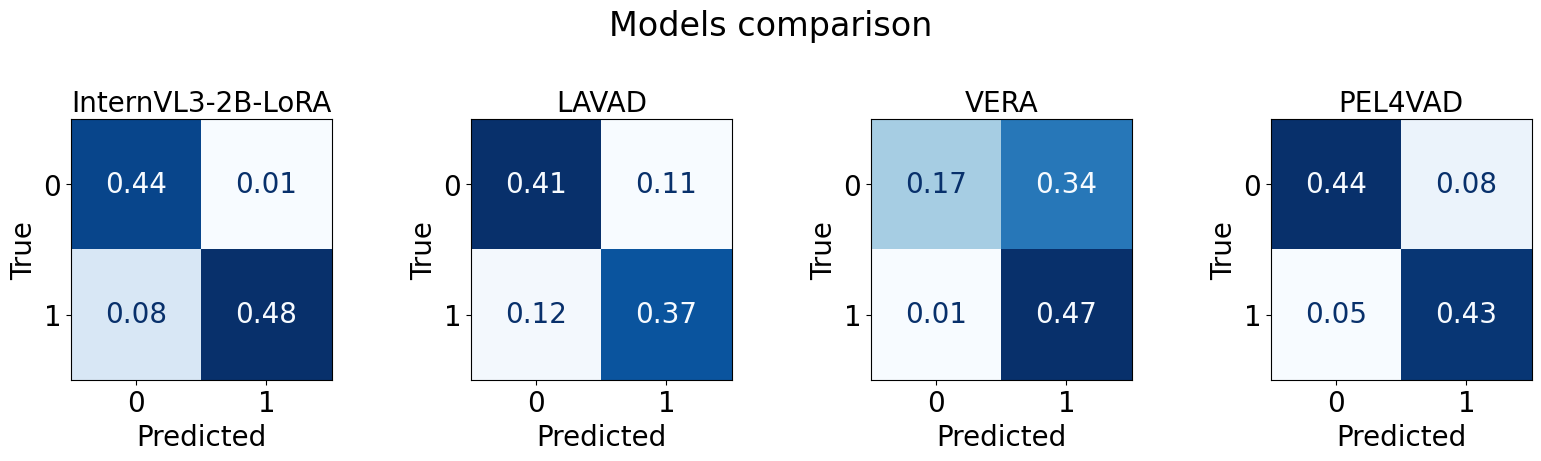}
\caption{{{Confusion} matrix for different models on UCF-crime dataset.}\label{fig:other_models}}
\end{figure}

Figure~\ref{fig:other_models} presents normalized confusion matrices for \texttt{InternVL3-2B}~\citep{zhu2025internvl3} finetuned with LoRA~\citep{hu2022lowrank}, LAVAD~\citep{Zanella2024HarnessingLL}, VERA~\citep{Ye2025VERA}, and PEL4VAD~\citep{pu2024learning} on the UCF-Crime~\citep{ucf_crime}, with rows denoting ground-truth classes and columns denoting predicted classes.

\vspace{-3pt}

\subsection{Illustrative Example}

\begin{figure}[H]
\includegraphics[width=13.86cm]{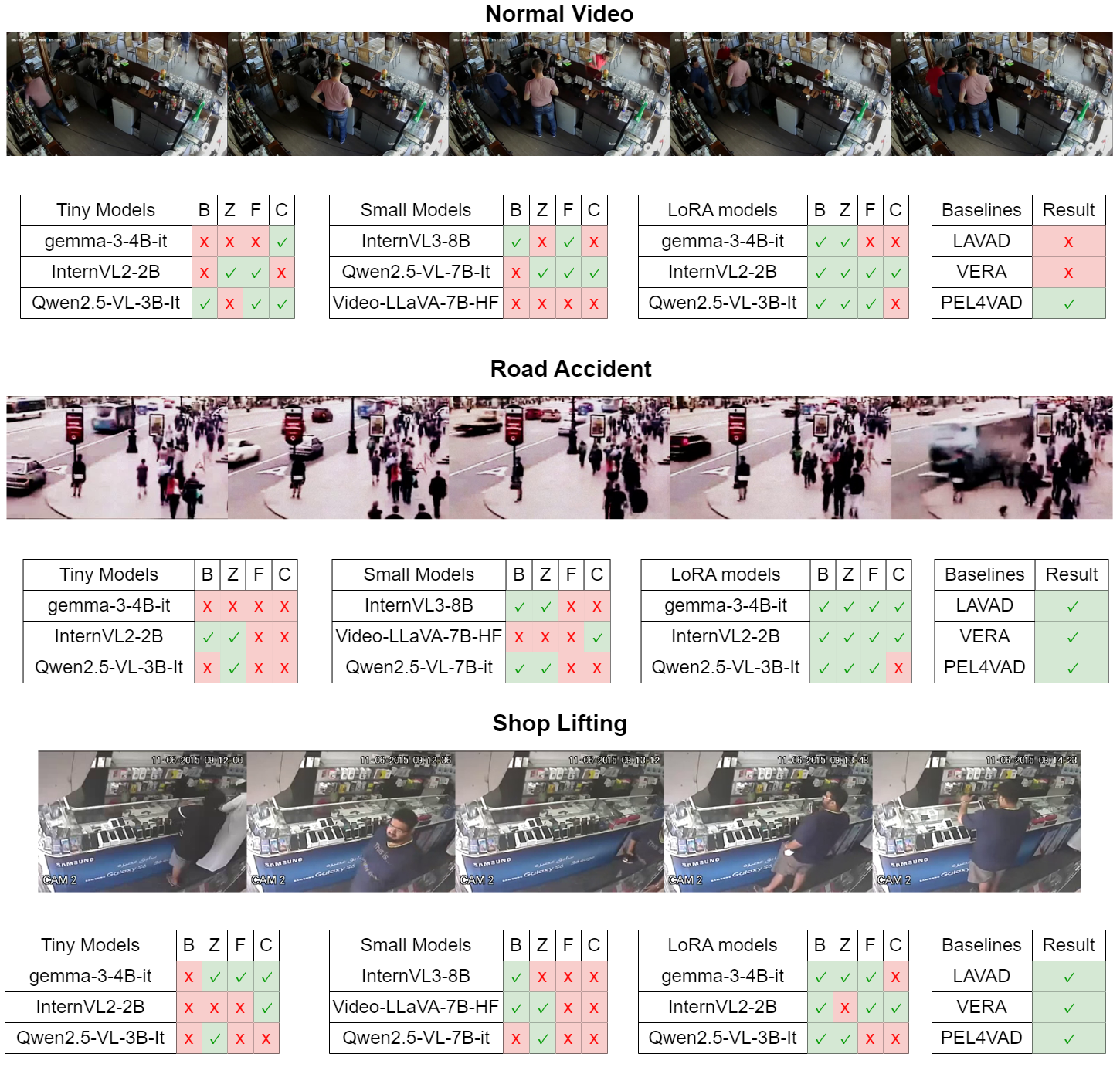}
\caption{{{Qualitative} clip‑level comparisons across models, prompts, and LoRA adaptation on representative surveillance scenarios.}\label{fig:illustrative}}
\end{figure}

{Figure~\ref{fig:illustrative} represents qualitative clip-level classification outcomes across compact vision--language models, prompts, and adaptation settings on three representative surveillance scenarios: Normal Video (benign activity), Road Accident (anomalous), and Shoplifting (anomalous). Each row group lists model families evaluated as Tiny Models, Small Models, and LoRA‑adapted variants, alongside classical baselines (LAVAD~\citep{Li2024VLAVAD}, VERA~\citep{Ye2025VERA}, PEL4VAD~\citep{pu2024learning}). Columns B, Z, F, and C denote Basic prompt, Zero‑shot prompt, Few‑shot prompt, and Chain‑of‑Thought prompt, respectively. {A green checkmark indicates a correct clip‑level decision relative to the scenario ground truth, while a red “x” indicates an incorrect decision.} This figure provides a side‑by‑side visualization of prompt sensitivity, model capacity effects, and the impact of lightweight LoRA adaptation~\citep{hu2022lowrank} on practical surveillance categories within the shared evaluation setting.}

\section{Discussion}

\subsection{Prompt Complexity Negatively Impacts Tiny and Small Models (``Overprompting'')}

\textls[-12]{Prompt complexity degrades performance for compact VLMs in this study: few-shot and chain-of-thought prompts consistently reduced F1 and accuracy while increasing latency, evidencing overprompting in both tiny and small tiers. Concise instruction-only or zero-shot prompts yielded more stable results for untuned models under surveillance~constraints.}

In the tiny tier, richer prompts harmed all three models despite identical evaluation conditions: \texttt{gemma-3-4B-it}~\citep{gemma3_2025} fell from F1 0.8182 with a Basic instruction to 0.6842 under CoT, alongside a latency jump from 9.10~s to {15.17~s} (\ref{tab:tiny_models_ucf_crime}). \texttt{Qwen2.5-VL-3B-It}~\citep{bai2025qwen2_5_vl} collapsed under few-shot prompting to F1 0.1818, and \texttt{InternVL3-2B}~\citep{zhu2025internvl3} deteriorated even further to F1 0.0142, indicating that longer, example-heavy schemas overwhelm limited-capacity models rather than help them ground the decision {boundary} (\ref{tab:tiny_models_ucf_crime}). Similar patterns appear in latency: \texttt{InternVL3-2B}'s~\citep{zhu2025internvl3} CoT inference expanded to 11.36~s versus 2.70~s for Basic, underscoring that added reasoning instructions impose nontrivial throughput costs without commensurate accuracy gains in the untuned {setting} (\ref{tab:tiny_models_ucf_crime}).

\subsection{Parameter-Efficient Fine-Tuning Demonstrates Transformative Capabilities}

Parameter-efficient fine-tuning with LoRA~\citep{hu2022lowrank} converted compact VLMs into high-fidelity detectors under the same protocol, yielding large F1 gains with equal or lower per-clip latency suitable for surveillance constraints. Across tiny models, these improvements consistently narrowed the gap relative to training-free use while preserving deployable throughput on UCF-Crime~\citep{ucf_crime}.

For example, \texttt{Qwen2.5-VL-3B-It}~\citep{bai2025qwen2_5_vl} rose from an F1 of 0.1818 in the few-shot regime to about 0.90 after LoRA~\citep{hu2022lowrank}, while \texttt{InternVL3-2B}~\citep{zhu2025internvl3} improved from 0.0142 to roughly~0.889, converting underperforming behaviors into reliable clip-level decisions (Table~\ref{tab:tiny_models_ucf_crime_lora} vs. Table~\ref{tab:tiny_models_ucf_crime}). Latency trends were favorable or neutral in most cases, such as \texttt{Qwen2.5-VL-3B-It}~\citep{bai2025qwen2_5_vl} (Basic) from 5.09~s to 2.83~s and \texttt{InternVL3-2B}~\citep{zhu2025internvl3} (CoT) from 11.36~s to about 5.09~s, strengthening the accuracy-latency trade-off.

\subsection{Parameter-Efficient Fine-Tuning Enhances Prompt Robustness}

Parameter-efficient fine-tuning with LoRA~\citep{hu2022lowrank} improved robustness across prompt types for compact VLMs on UCF-Crime~\citep{ucf_crime}, yielding consistently higher F1 and ROC-AUC under Basic, Zero-Shot, Few-Shot, and CoT prompts. These gains reflect stable performance across instruction regimes rather than dependence on any single prompting style (Table~\ref{tab:tiny_models_ucf_crime_lora}).

After LoRA~\citep{hu2022lowrank}, \texttt{gemma-3-4B-it}~\citep{gemma3_2025} produced tightly clustered F1 scores of 0.91--0.92 across Basic, Zero-Shot, Few-Shot, and CoT with +11--33\% improvements versus non-LoRA, evidencing prompt-invariant behavior at the detector level (Table~\ref{tab:tiny_models_ucf_crime_lora}). Similar patterns hold for \texttt{Qwen2.5-VL-3B-It}~\citep{bai2025qwen2_5_vl} and \texttt{InternVL3-2B}~\citep{zhu2025internvl3}, indicating that fine-tuning normalizes outcomes across prompting styles while also reducing or holding latency steady in many cases, e.g., \texttt{InternVL3-2B}~\citep{zhu2025internvl3} CoT at 5.09~s and \texttt{gemma-3-4B-it}~\citep{gemma3_2025} CoT at 8.77 s, both reported with favorable percentage shifts relative to non-LoRA.

\subsection{Specialized Compact Models Outperforms General-Purpose Larger Counterparts}

Specialized compact VLMs adapted with parameter-efficient fine-tuning outperformed several general-purpose 7-8B baselines on UCF-Crime~\citep{ucf_crime} in F1 while offering competitive or lower per-clip latency, yielding a superior accuracy-efficiency trade-off for surveillance deployment.

Under the same protocol, LoRA-tuned~\citep{hu2022lowrank} \texttt{gemma-3-4B-it}~\citep{gemma3_2025} attains F1 of 0.91 with Basic prompting, exceeding \texttt{Qwen2.5-VL-7B-It}~\citep{bai2025qwen2_5_vl} at 0.9010 and \texttt{InternVL3-8B}~\citep{zhu2025internvl3} at 0.8538, while compact LoRA~\citep{hu2022lowrank} models often run faster {(e.g., 2.83–2.91 s)} than 7-8B baselines {(3.71–4.89 s)} (Tables~\ref{tab:small_models_ucf_crime},~\ref{tab:tiny_models_ucf_crime_lora}). Similar trends hold across other prompts, where compact LoRA~\citep{hu2022lowrank} models achieve higher F1 than 7-8B baselines with competitive latency on UCF-Crime~\citep{ucf_crime}.

\subsection{Simplified Architectures Achieve Competitive Performance}

\textls[-15]{Streamlined {architectures} with parameter-efficient adaptation achieve competitive classification quality under a unified clip-level protocol on UCF-Crime~\citep{ucf_crime}, and the proposed approach demonstrates competitive latency when evaluated alongside established~baselines.}

The method maintains simplicity of implementation and reproducibility while operating under the same experimental conditions as comparative systems, indicating a balanced accuracy-efficiency profile without introducing additional architectural stages.

\subsection{Task-Specific Adaptation Is Essential for Effective Anomaly Detection}

Task-specific adaptation via parameter-efficient fine-tuning is essential for effective anomaly detection, delivering consistent gains over training-free use across models and prompt types on UCF-Crime~\citep{ucf_crime}. Within the same protocol, lightweight LoRA~\citep{hu2022lowrank} adapters raise F1 and ROC-AUC, showing that domain alignment is required to achieve reliable detection quality rather than optional fine-tuning.

Adaptation aligns decision boundaries with surveillance semantics and dataset priors, improving precision-recall balance and enabling stable operating-point selection without dependence on particular prompt formulations. These effects hold across compact VLM architectures under identical splits and evaluation criteria, indicating that task-specific tuning is a prerequisite for robust anomaly detection in this setting.

\subsection{Qualitative Results}

{LoRA adaptation consistently strengthened compact VLMs by converting unstable training‑free behaviors into reliable clip‑level decisions while also reducing prompt sensitivity and, in several cases, lowering per‑clip latency under the unified protocol (Figure~\ref{fig:illustrative}). Concretely, LoRA raised accuracy and F1 across Basic, Zero‑shot, Few‑shot, and CoT prompts for tiny models, aligning decision boundaries with surveillance semantics and yielding prompt‑invariant performance patterns without increasing architectural complexity. These qualitative gains are evident in Figure~\ref{fig:illustrative}, where LoRA~\citep{hu2022lowrank} rows produce more checkmarks across both anomalous clips (Road Accident, Shoplifting) and the benign clip (Normal Video), indicating improved precision on normal activity and recall on subtle anomaly cues under identical inputs and thresholds.}

\subsection{Limitations}

This study operates at the clip level with weak supervision and RGB-only inputs using fixed sampling, which can miss brief audio- or motion-centric cues that are outside scope here. Interpretability is constrained by the absence of temporal localization, but a practical mitigation is to elicit rationale via conversational VLM queries to surface decisive frames and factors when needed. 

The evaluation is single-camera by design and does not include multi-camera fusion or cross-camera reasoning, aligning with prevalent benchmarks that provide clip-level single-view footage rather than synchronized multi-view datasets. Parameter-efficient fine-tuning is instantiated with a fixed LoRA~\citep{hu2022lowrank} recipe; while broader PEFT variants were not explored, evidence indicates that, when applied across layers with adequate rank, LoRA can match full fine-tuning on supervised tasks~\citep{schulman2025lora}, supporting the chosen adaptation strategy in this setting.

{Large-scale vision--language models adapted with LoRA~\citep{hu2022lowrank} present additional constraints that differ from those observed in compact configurations. As model capacity increases, adapter placement, rank selection, and training stability become more sensitive to initialization and optimization settings, often requiring case-specific tuning to achieve convergence without overfitting. Furthermore, emergent behaviors in large models---such as complex prompt dependencies and activation saturation---may limit LoRA's ability to fully align task-specific decision boundaries without revisiting adapter distribution or expanding rank budgets. These factors indicate that while LoRA enables efficient adaptation in compact models, its scalability to very large architectures is not guaranteed, and systematic evaluation across depth, capacity, and parameter budget remains necessary to establish reliable performance scaling under parameter-efficient fine-tuning.}

\subsection{Future Work}

{Future work will extend toward real-time deployment scenarios and multimodal data integration to further align model design with operational surveillance demands. Real-time adaptation requires optimizing inference pipelines for low-latency execution on edge or embedded hardware while maintaining stable accuracy across variable frame rates and environmental conditions. Incorporating multimodal cues---such as audio offers a pathway to enhance anomaly discrimination and robustness under occlusion, noise, or visually ambiguous events. Integrating these modalities through lightweight fusion or cross-attention mechanisms compatible with parameter-efficient fine-tuning remains an open direction, aiming to preserve the compactness and transparency of the current framework while broadening situational awareness in realistic, resource-constrained deployments.}

\section{Conclusions}

Parameter-efficient fine-tuning emerges as the critical enabler for compact vision-language models in video anomaly detection, turning training-free variants into reliable clip-level detectors while retaining competitive per-clip latency within the shared protocol. After adaptation, concise instruction or zero-shot prompting remains sufficient, and prompt sensitivity is notably reduced across models and settings, yielding consistent behavior under the same evaluation conditions.

When compared under identical splits, metrics, and runtime settings, the adapted compact models perform on par with, and in multiple cases exceed, established approaches representative of training-free VLM pipelines and weakly supervised MIL-style baselines, while maintaining comparable or lower inference time per clip. These findings position parameter-efficiently adapted compact VLMs as strong contenders among contemporary alternatives evaluated in this study, offering a favorable balance of accuracy and latency without increasing architectural complexity.

\vspace{6pt}

\authorcontributions{Conceptualization, G.M. and M.G.; methodology, K.B.; software, K.B.; validation, K.B., K.K., N.V., K.G. and I.L.; formal analysis, K.B. and G.M.; investigation, K.B. and G.M.; resources, K.B., G.M., M.G., K.K. and K.G.; data curation, K.B.; writing---original draft preparation, K.B. and G.M.; writing---review and editing, K.B. and G.M.; visualization, K.B.; supervision, G.M. and M.G.; project administration, G.M. and M.G.; funding acquisition, G.M. and M.G. All authors have read and agreed to the published version of the manuscript.}

\funding{{This} research received no external funding.}

\institutionalreview{Not applicable.}

\informedconsent{Not applicable.}

\dataavailability{
\textls[-15]{\texttt{Qwen2.5-VL} \{collection}~\citep{bai2025qwen2_5_vl}---\url{https://huggingface.co/collections/Qwen/qwen25-vl-6795ffac22b334a837c0f9a5}; 
LAVAD~\citep{Zanella2024HarnessingLL}---\url{Zanella2024HarnessingLL}; 
\texttt{Video-LLaVA-7b}} \citep{lin2024videollava}---\url{https://huggingface.co/LanguageBind/Video-LLaVA-7B-hf};
PEL4VAD~\citep{pu2024learning}---\url{https://github.com/yujiangpu20/PEL4VAD}; 
\texttt{Gemma-3} collection~\citep{gemma3_2025}---\url{https://huggingface.co/collections/google/gemma-3-release-67c6c6f89c4f76621268bb6d}; 
Project repository---\url{https://github.com/KORALLLL/DeviantBehaviorResearch/tree/kirill};
\texttt{InternVL3} collection~\citep{zhu2025internvl3}---\url{https://huggingface.co/collections/OpenGVLab/internvl3-67f7f690be79c2fe9d74fe9d}; 
VERA~\citep{Ye2025VERA}---\url{https://github.com/vera-framework/VERA}; 
UCF-crime dataset~\citep{ucf_crime}---\url{https://huggingface.co/datasets/jinmang2/ucf_crime}; 
LLaMA factory project~\citep{zheng2024llamafactory}---\url{https://github.com/hiyouga/LLaMA-Factory}.

\conflictsofinterest{The authors declare no conflicts of interest.} 


\newpage

\abbreviations{Abbreviations}{
The following abbreviations are used in this manuscript:
\\

\noindent 
\begin{tabular}{@{}ll}
VERA & Verbalized learning framework~\citep{Ye2025VERA}\\
FPR & False positive rate\\
VLM & Vision-language model\\
TEF-VAD & Transformer Encoded Feature Video Anomaly Detection~\citep{10.1007/978-981-96-2641-0_8}\\
LLM & Large language model\\
MIL & Multiple instance learning\\
OT-CTST & Outlier-Embedded Cross Temporal Scale Transformer~\citep{majhi2024oectst}\\
RTFM & Robust Temporal Feature Magnitude Learning~\citep{Tian2021WeaklySupervised}\\
LoRA & Low-Rank Adaptation~\citep{hu2022lowrank}\\
CCTV & Closed-circuit television\\
PEFT & Parameter-efficient fine-tuning\\
TPR & True positive rate\\
VAD & Video Anomaly Detection\\
DSS-Net & Dynamic self-supervised network~\citep{10.1109/TMM.2023.3292596}\\
LAVAD & Language-based Video Anomaly Detection~\citep{Zanella2024HarnessingLL}\\
RGB & Red-Green-Blue color model\\
ROC AUC & Area Under the Receiver Operating Characteristic Curve\\
WSVAD & Weakly Supervised Video Anomaly Detection~\citep{Lv2023Unbiased}\\
\end{tabular}
}

\appendixtitles{yes} 
\appendixstart
\appendix
\section[\appendixname~\thesection]{Prompts}\label{asec:prompts}

\subsection[\appendixname~\thesubsection]{Basic Prompt}\label{asubsec:basic_prompt}

\noindent\fbox{
  \begin{minipage}{0.95\textwidth}
    {Return} 
 ‘1‘ if the video shows any deviant, abnormal or criminal behavior; return ‘0‘ if it does not. Respond with only
that single digit and nothing else.
  \end{minipage}
}

\subsection[\appendixname~\thesubsection]{Zero-Shot Prompt}\label{asubsec:zeroshot_prompt}

\noindent\fbox{
  \begin{minipage}{0.95\linewidth}
You are a multimodal vision--language model.\\
Determine whether the video exhibits deviant, abnormal, or criminal behavior.\\\\
INTERNAL STEPS\\
1. Observe the most salient actions, objects, and spoken words.\\
2. Apply your own knowledge of what constitutes deviant behavior.\\
3. Decide if the content is deviant (positive) or not (negative).\\\\
OUTPUT\\
Return only one character:\\
• ‘1‘ → deviant detected\\
• ‘0‘ → no deviance detected\\\\
Do **not** reveal your reasoning or the internal steps---reply with the single digit only
  \end{minipage}
}

\subsection[\appendixname~\thesubsection]{Few-Shot Prompt}\label{asubsec:fewshot_prompt}
\noindent\fbox{
  \begin{minipage}{0.95\linewidth}
ROLE\\
You are an advanced vision--language model.\\
Your sole goal is to watch a short video clip and decide whether it contains any clearly visible act of deviant, abnormal, or criminal behavior.\\\\
A clip is POSITIVE (‘1‘) if at least one of the following categories is observed with high confidence; otherwise it is NEGATIVE (‘0‘):\\
• Abuse---non-consensual harming of a vulnerable victim\\
• Fighting---two or more people mutually exchanging
violence\\
• Car accident---traffic collision or hit-and-run incident\\
• Explosion/Fire/Arson---deliberate or accidental ignition causing damage\\
• Shooting---discharge of a firearm toward a target\\
• Riot/Crowd violence---large-scale disorder with property destruction or assault\\\\
INTERNAL REASONING (not to be shown)\\
1. Visually examine key frames (motion, objects, people, context).\\
2. Compare observations against the six categories.\\
3. If any category matches, output ‘1‘; otherwise output ‘0‘.\\
4. Do **not** reveal your reasoning---return only the digit.\\
==================== FEW-SHOT EXAMPLES ====================\\
The 14 few-shot examples are released as
prompts//fewshot.txt in the project repository\\
==========================================================\\\\
YOUR ANSWER\\
Return **only one character** on a single line:\\
• ‘1‘---deviant behavior detected\\
• ‘0‘---no deviant behavior\\

  \end{minipage}
}

\subsection[\appendixname~\thesubsection]{Chain-of-Thought Prompt}\label{asubsec:cot_prompt}

\noindent\fbox{
  \begin{minipage}{0.95\linewidth}
ROLE\\
You are an advanced vision--language model.\\
Your task is to inspect a short video clip and decide whether it contains any clear act of deviant, abnormal, or criminal behavior.\\
A clip is POSITIVE (1) if at least one of the following categories is observed with high confidence; otherwise it is NEGATIVE (0):\\
• Abuse---non-consensual harming of a vulnerable victim\\
• Fighting---two or more people mutually exchanging
violence\\
• Car accident---traffic collision or hit-and-run incident\\
• Explosion/Fire/Arson---deliberate or accidental ignition
causing damage\\
• Shooting---discharge of a firearm toward a target\\
• Riot/Crowd violence---large-scale disorder with property destruction or assault\\\\
 \end{minipage}
}

\noindent\fbox{
  \begin{minipage}{0.95\linewidth}
THINKING PROTOCOL\\
1. **Observe**---Describe salient visual elements (people, actions, objects, motion, context).\\
2. **Reason**---Map observations to any deviance category; note level of confidence.\\
3. **Double-check**---Rethink the scene for missed cues or false positives.\\
4. **Decide**---Output 1 if any category matches; else 0.\\
5. **Respond**---Show the full chain of thought for steps 1-4, then put the single digit **alone** on the final line.\\
==================== FEW-SHOT EXAMPLES ====================\\
The 7 few-shot examples are released as prompts//cot.txt in the project repository\\
==========================================================\\\\
NOW FOLLOW THE THINKING PROTOCOL AND RETURN THE SINGLE-DIGIT ANSWER ON THE FINAL LINE.
  \end{minipage}
}


\begin{adjustwidth}{-\extralength}{0cm}
\reftitle{References}



\PublishersNote{}
\end{adjustwidth}

\end{document}